\documentclass[letterpaper, 10 pt, conference]{ieeeconf}
\IEEEoverridecommandlockouts
\let\labelindent\relax

\usepackage{cite}
\usepackage{amsmath,amssymb,amsfonts}
\usepackage[ruled,linesnumbered,boxed,noend]{algorithm2e}
\usepackage{graphicx}
\usepackage{textcomp}
\usepackage{xcolor}
\usepackage{booktabs}
\usepackage{tabularx} 
\usepackage{multirow} 
\usepackage{threeparttable}
\usepackage{enumitem}
\usepackage[font={small}]{caption}
\usepackage{balance}
\usepackage{textcomp}
\usepackage{stfloats}
\usepackage{siunitx}
\usepackage{bm}
\usepackage{epstopdf}
\usepackage[colorlinks, linkcolor=black, anchorcolor=red, citecolor=black]{hyperref}

\title{\LARGE \bf 
GGD-SLAM: Monocular 3DGS SLAM Powered by Generalizable Motion Model for Dynamic Environments
}

\author{ Yi Liu, Haoxuan Xu, Hongbo Duan, Keyu Fan, Zhengyang Zhang, \\ Peiyu Zhuang, Pengting Luo, Houde Liu$^{*}$
\thanks{This work was supported by the Shenzhen Science and Technology Program (Grant No. RCJC20210706091946001) and the Shenzhen Science and Technology Program (Grant No. ZDCY20250901104207008).}
\thanks{Yi Liu, Hongbo Duan, Keyu Fan, Zhengyang Zhang and Houde Liu are with Shenzhen International Graduate School, Tsinghua University, Shenzhen, China.}
\thanks{Haoxuan Xu is with Thrust of Robotics and Autonomous Systems, The Hong Kong University of Science and Technology (Guangzhou), Guangzhou, China.}
\thanks{Peiyu Zhuang is with the School of Cyber Science and Technology, Sun Yat-sen University, Shenzhen, China.}
\thanks{Pengting Luo is with the Central Media Technology Institute, Huawei Incorporated
Company, Shenzhen, China.}
\thanks{* Corresponding authors: Houde Liu (Emails: liu.hd@sz.tsinghua.edu.cn) }
}

\begin{document}
\maketitle
\begin{abstract}
Visual SLAM algorithms achieve significant improvements through the exploration of 3D Gaussian Splatting (3DGS) representations, particularly in generating high-fidelity dense maps. However, they depend on a static environment assumption and experience significant performance degradation in dynamic environments. This paper presents GGD-SLAM, a framework that employs a generalizable motion model to address the challenges of localization and dense mapping in dynamic environments—without predefined semantic annotations or depth input. Specifically, the proposed system employs a First-In-First-Out (FIFO) queue to manage incoming frames, facilitating dynamic semantic feature extraction through a sequential attention mechanism. This is integrated with a dynamic feature enhancer to separate static and dynamic components. Additionally, to minimize dynamic distractors' impact on the static components, we devise a method to fill occluded areas via static information sampling and design a distractor-adaptive Structure Similarity Index Measure (SSIM) loss tailored for dynamic environments, significantly enhancing the system's resilience. Experiments conducted on real-world dynamic datasets demonstrate that the proposed system achieves state-of-the-art performance in camera pose estimation and dense reconstruction in dynamic scenes.
\end{abstract}


\section{Introduction}
\label{sec:Introduction}
Visual simultaneous localization and mapping (vSLAM) \textemdash the task of estimating the pose of a vision sensor and reconstructing a static map of an unknown environment\textemdash is a core technology for mobile robotics, augmented reality (AR) and virtual reality (VR)\cite{2}. Furthermore, compared to sparse reconstruction, dense reconstruction provides superior geometric precision and detail fidelity, enabling a more comprehensive and accurate representation of the scene geometry and texture\cite{3}. However, mobile robots often operate in dynamic environments where unpredictable changes can cause dense SLAM algorithms to suffer from significant performance degradation\cite{Gassidy}.

\begin{figure}[t]
    \centering
    \includegraphics[width=8.5cm]{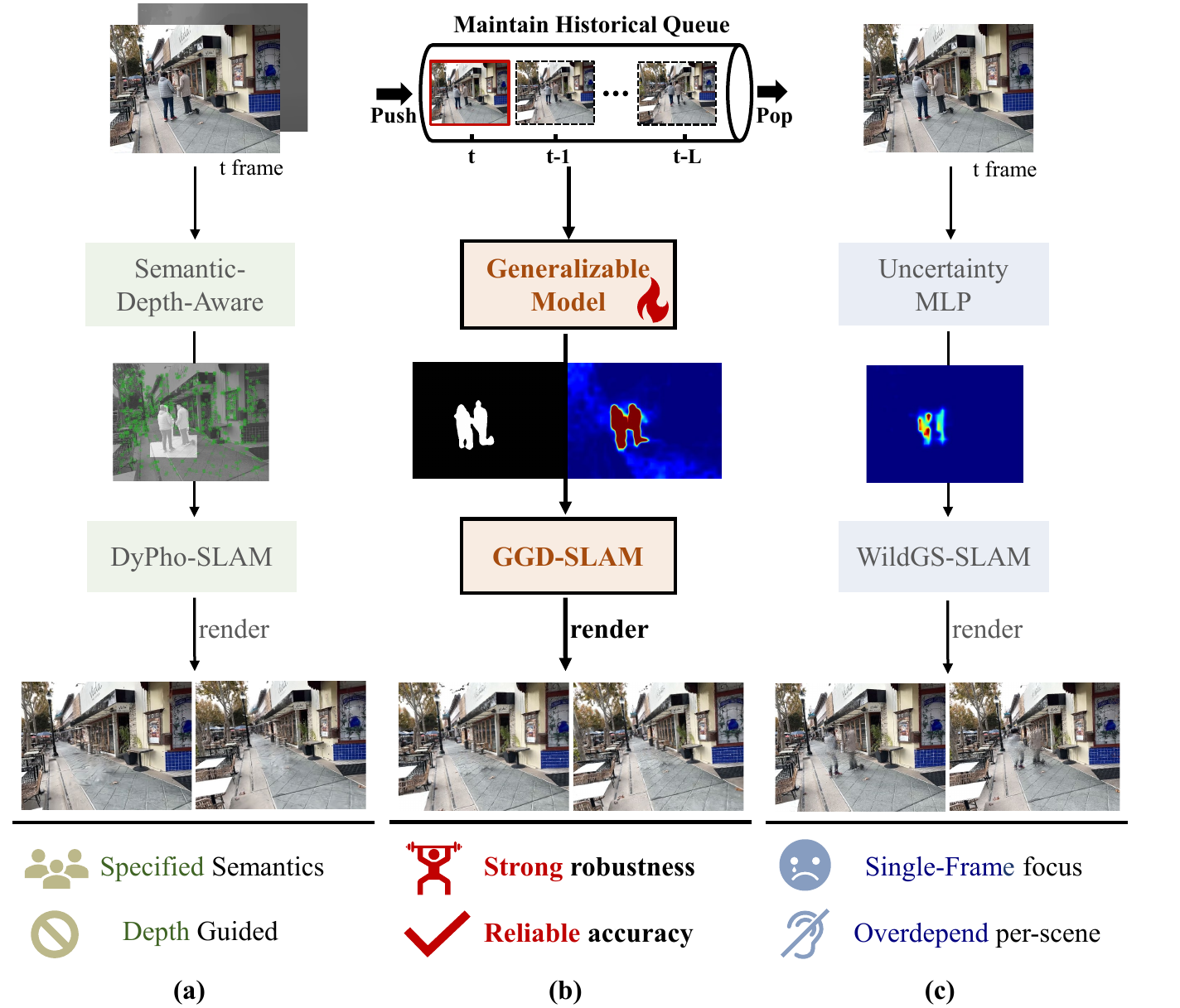}
    \caption{\textbf{Motivation}: Left: DyPho-SLAM\cite{dypho-slam} requires specific semantic labels and depth input for dynamic removal; Right: The MLP in WildGS-SLAM\cite{wildgs} is constrained by per-scene rendering performance; Our GGD-SLAM introduces a generalizable motion model without semantic labels or depth input, reducing the need for per-scene 3DGS rendering loss supervision.}
    \label{first}
    \vspace{-0.5cm}
\end{figure}

Accomplishing dense SLAM tasks in dynamic environments presents the main difficulty: Without predefined semantic labels, it is challenging to identify dynamic semantics from the incremental input images of SLAM \cite{addslam}, leading to the occurrence of motion artifacts. In Fig. \ref{first}(a), DyPho-SLAM \cite{dypho-slam} requires specific predefined labels and depth input to remove dynamic feature points. In Fig. \ref{first}(c), the latest approach WildGS-SLAM\cite{wildgs} introduces an uncertainty-aware approach that employs an MLP supervised by 3DGS rendering loss to identify the dynamic region. Nevertheless, this approach heavily depends on the quality of 3DGS per-scene rendering. In areas where 3DGS reconstruction is weak (e.g., background areas) or where dynamic objects are transiently static (e.g., wandering person in stop-and-go motion in Wild-SLAM/Wandering dataset as Fig. \ref{first}), it can lead to severe misjudgment of dynamic objects\cite{wildgs}, thereby degrading the overall reconstruction quality.

 To mitigate these issues, a critical insight needs to be considered: dynamic object identification across temporal sequences. In the context of SLAM, where image frames are processed progressively, a single frame can only provide static attributes of objects, such as contour, shape, texture, and semantic labels\cite{sam}, but lacks any explicit motion information. The motion nature is inherently defined by its positional change over a sequence of frames. Accurate dynamic identification requires the incorporation of temporal context and correlation across multiple consecutive frames.

Through a detailed analysis of the above methods, this paper proposes a monocular 3D\textbf{G}S SLAM system powered by \textbf{G}eneralizable motion model for \textbf{D}ynamic environments, named GGD-SLAM as shown in Fig. \ref{first}(b). To be specific, we first introduce a First-In-First-Out (FIFO) queue structure to manage progressively incoming frames in the SLAM system, which facilitates dynamic semantic feature extraction via a sequential attention mechanism and is integrated with a dedicated dynamic feature enhancer for separating dynamic and static components. Next, we propose a distractor-adaptive  Structure Similarity Index Measure (SSIM) loss to effectively minimize the impact of dynamic distractors on the static components. Then, we construct a KD-tree for the static Gaussians and fill in occluded areas by sampling static information, thereby preserving the consistency of the static background. Finally, we integrate the generalizable model with an uncertainty-aware approach, reducing misjudgment of dynamic objects caused by sole reliance on 3DGS rendering errors, thereby enabling precise tracking and high-fidelity mapping in dynamic settings.

Our primary contributions are as follows:
\begin{enumerate}

\item We present a generalizable motion model (GMM) to extract dynamic semantics for progressive SLAM inputs, which enables high-quality dense mapping and tracking in real-world dynamic environments without relying on predefined semantic annotations or depth information.
\item We propose a mapping optimization strategy for dynamic scene reconstruction, which employs a static Gaussian KD-tree to preserve background consistency and incorporates a distractor-adaptive SSIM method for dynamic scenarios, thereby enabling accurate photorealistic reconstruction.
\item Experiments on a wide variety of challenging real-world dynamic datasets demonstrate that GGD-SLAM achieves state-of-the-art performance in camera pose estimation and dense map reconstruction.

\end{enumerate}

\section{Related Work}
\subsection{Traditional Dynamic SLAM}
Current research on dynamic SLAM primarily relies on single-frame object detection, neglecting the intrinsic nature of object motion across frames, and thus fails to achieve robust extraction consistently across all frames. For example, ORB-SLAM3 \cite{orbslam} adopts re-sampling and residual optimization strategies to remove dynamic objects. Dyna-SLAM \cite{dynaslam} utilizes semantic extraction to generate dynamic object masks, which limits its generalizability in real-world scenarios. Rodyn-SLAM \cite{rodyn-slam} further integrates optical flow-based motion estimation on this basis; however, the optical flow method relies merely on image information from neighboring frames of short duration and often suffers from misjudgment when faced with complex motion conditions. \par
\subsection{Dynamic 3DGS SLAM}
Considering the resource consumption issue of implicit representations \cite{5}, 3DGS meets the requirement for fast rendering through explicit scene representations, thus facilitating its rapid development in the field of dense SLAM \cite{10}. While MonoGS \cite{gsslam} and SplatAM \cite{splatam} perform well in tracking and reconstruction, their performance degrades significantly in dynamic environments. DG-SLAM\cite{dgslam} attempts to remove dynamic objects using a depth warp approach, but cannot yet overcome the problem of misjudgment in dynamic extraction that relies on neighboring frames of short duration. DyPho-SLAM\cite{dypho-slam} employs prior depth information to effectively track and eliminate dynamic objects; however, same as UP-SLAM\cite{upslam} and DGS-SLAM\cite{dgs-slam}, it requires absolute depth inputs. The latest approach WildGS-SLAM\cite{wildgs} uses a simple uncertainty-aware MLP to identify dynamic regions, but it relies heavily on 3DGS rendering—its performance deteriorates significantly in regions with rendering artifacts. To address this dynamic removal challenge, we propose a generalizable dynamic extractor that leverages attention mechanisms on a queue of historical frames to extract dynamic semantics, without relying on predefined semantic annotations or depth input.
\subsection{Moving Object Segmentation Method}
Beyond the field of dynamic 3DGS SLAM, moving object segmentation (MOS)\cite{mos} technology is frequently employed in domains such as video processing and feedforward reconstruction. MegaSAM\cite{megasam} only leverages neighboring frames and lacks sufficient multi-view information, leading to less reliable motion masks. MonST3R\cite{monst3r} generates masks by comparing the actual input image with a dynamically deformed canonical point cloud; however, this differs from the task formulation of dynamic 3DGS SLAM for static regions and cannot be applied to 3D Gaussian reconstruction. Other video-based dynamic object extraction works, such as Segment-Any-Motion\cite{segment}, utilize long-range tracking with SAM2 to enable efficient dynamic mask densification, yet they cannot be well adapted to SLAM systems with progressive inputs.

\begin{figure*}[t]
    \centering
    \includegraphics[width=18cm]{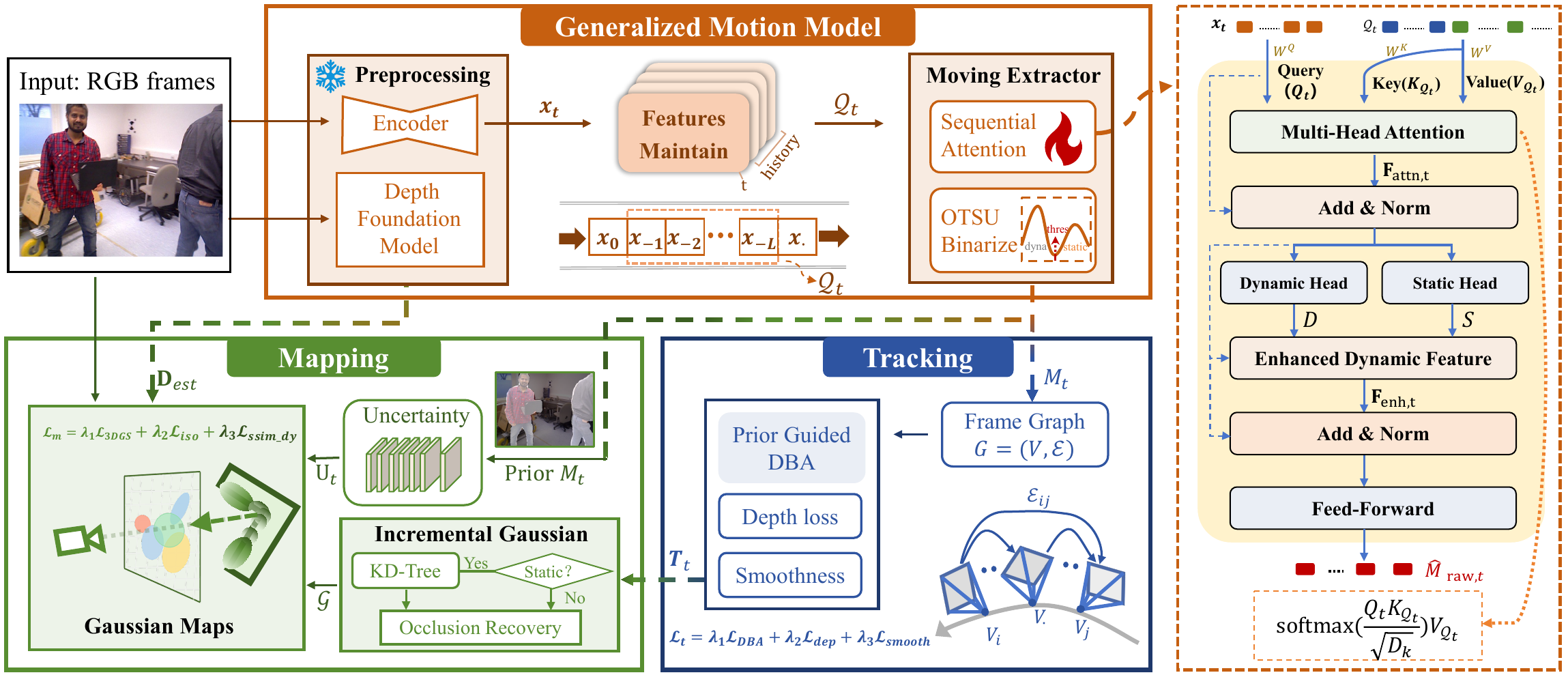}
    \caption{Pipeline of our GGD-SLAM: the ``Generalized Motion Model" (\ref{subsec:NLOS Identification}) processes progressive frame $\boldsymbol{I}_t$ to extract prior $M_t$, followed by encoder preprocessing, historical features $\mathcal{Q}_{t}$ maintenance, and the moving extractor. Subsequently, the ``Tracking" module (\ref{subsec:Location Algorithm}) passes through the frame graph $\mathcal{E}$ maintenance, and DBA optimization guided by prior to obtain the camera pose $\boldsymbol{T}_t$. Finally, the ``Mapping" module (\ref{subsec:Mapping Algorithm}) utilizes priors to train the uncertainty model, and incrementally adds Gaussians as well as optimizes the Gaussian map $\mathcal{G}$. }
    \label{Fig:framework}
    \vspace{-0.5cm}
\end{figure*}
\section{Implementation}
\label{sec:implementation}
Given an image sequence $\mathcal{I}=\{\boldsymbol{I}_i\}_{i=1}^N (\boldsymbol{I}_i\in \mathbb{R}^{H\times W\times3})$ containing dynamic objects captured at a certain frame rate by a monocular camera with known intrinsics, our objective is to obtain their corresponding camera pose matrices $\mathcal{T}=\{\boldsymbol{T}_i\}_{i=1}^N$, while updating Gaussian parameters $\mathcal{G}=\{\boldsymbol{\mu}_j,\alpha_j,\boldsymbol{\Sigma_j},\boldsymbol{c}_j\}_{j=1}^{n(\mathcal{G})}$in every frame. The Gaussian parameter is characterized by its spatial position $\boldsymbol{\mu}_j$, opacity value $\alpha_j$, covariance matrix $\boldsymbol{\Sigma}_j$, and spherical harmonics-based color coefficients $\boldsymbol{c}_j$.
The process of solving $\mathcal{T}$ and $\mathcal{G}$ should satisfy:  
1) minimizing the absolute trajectory error root mean square error.
2) optimal Gaussian rendering performance for representing the static environment. The structure of the proposed GGD-SLAM algorithm completes localization and dense mapping tasks in dynamic environments through a series of interconnected processes, as shown in Fig. \ref{Fig:framework}.

\subsection{Generalized Motion Model}
\label{subsec:NLOS Identification}
Our main contribution lies in designing a generalizable motion model for a dynamic semantics extractor within image sequences, designed specifically for progressive SLAM systems as Algorithm \ref{alg:dyn-prior}. This generalizable motion model eliminates the need for per-scene online training, serving as a robust prior for GS-SLAM systems.

\subsubsection{Data Preprocessing}
\ 
\newline
\indent 
Given an input image $\boldsymbol{I}_t$, we employ a pre-trained DINOv2 feature extractor to obtain image features $\mathbf{x}_t=\text{DINOv2}(\boldsymbol{I}_t) \in \mathbb{R}^{H'\times W'\times C}$. Here, $H'$ and $W'$ denote the spatially downsampled dimensions of the feature map resulting from patch embedding, while $C$ denotes the channel dimension. However, $\mathbf{x}_t$ merely captures structural features (e.g., contours, shapes, textures) and semantic information from $\boldsymbol{I}_t$, which lacks temporal dynamics for modeling scene evolution. To address this limitation, we introduce a FIFO queue structure $\mathcal{Q}_{t}$ that progressively aggregates sequential frames for dynamic feature extraction. Specifically, each input feature vector $\mathbf{x}_t$ iteratively updates $\mathcal{Q}_{t}\in \mathbb{R}^{L\times H'\times W'\times C}$ via this mechanism:
\begin{equation}
\mathcal{Q}_{t} = 
\begin{cases} 
\left[ \mathbf{x}_{t-L},\ \mathbf{x}_{t-L+1},\ \ldots,\ \mathbf{x}_{t-1} \right] & t \geq L \quad \\[2mm]
\left[ \smash{\underbrace{\mathbf{0},\ \ldots,\ \mathbf{0}}_{L-t+1\text{ items}}} ,\ \mathbf{x}_1,\ \ldots,\ \mathbf{x}_{t-1}\right] & t < L \quad
\end{cases}
\vspace{0.3cm}
\end{equation}
For non-full queues, we prepend zero-padding vectors $\mathbf{0}$ to maintain temporal consistency. The dynamic semantics is determined by temporal variation across $L$ frames—a larger $L$ provides a broader temporal context for motion inference.

\subsubsection{Sequential Attention Mechanism}
\ 
\newline
\indent 
After obtaining the current structural feature $\mathbf{x}_t$ and historical features $\mathcal{Q}_t$, we integrate contextual information through an attention mechanism as follows:
\begin{subequations}
  \begin{align}
    \mathbf{Q}_t = \mathbf{x}_t \mathbf{W}^Q, \quad 
\mathbf{K}_t &= \mathcal{Q}_t \mathbf{W}^K, \quad 
\mathbf{V}_t = \mathcal{Q}_t \mathbf{W}^V\\
    \mathbf{F}_{\text{attn},t} =& \text{Attention}(\mathbf{Q}_t, \mathbf{K}_t, \mathbf{V}_t) 
  \end{align}
\end{subequations}
where $\mathbf{Q}_t\in \mathbb{R}^{(1\times H'\times W')\times C}$ is the query vector from the current frame's features. $\mathbf{K}_t \in \mathbb{R}^{(L\times H'\times W')\times C}$ and $\mathbf{V}_t \in \mathbb{R}^{(L\times H'\times W')\times C}$ establish a temporal retrieval index from historical features, enabling cross-frame matching. The output $\mathbf{F}_{\text{attn},t}\in \mathbb{R}^{H'\times W'\times C}$ encapsulates temporally-enhanced features for generalizable motion semantics segmentation.

To achieve enhanced dynamic-static separation, we process $\mathbf{F}_{\text{attn},t}$ through two separate heads, a dynamic head and a static head, obtaining disentangled representations: an enhancement coefficient $\mathbf{D}\in \mathbb{R}^{H'\times W'\times C}$ for dynamic attributes and a suppression coefficient $\mathbf{S}\in \mathbb{R}^{H'\times W'\times C}$ for static components. Subsequently, these representations are fused via a gated attention mechanism as follows: 
\begin{equation}
    \label{equ:distribution}
	\begin{aligned}
\mathbf{F}_{\mathrm{enh},t} = \mathbf{F}_{\text{attn},t} \odot \underbrace{\left( 1 + \alpha \mathbf{D} \right)}_{\text{Dynamic}} \odot \underbrace{\left( 1 - \alpha \mathbf{S} \right)}_{\text{Static}}
	\end{aligned}
\end{equation}
where $\odot$ denotes the Hadamard product, and the balancing coefficient $\alpha$ is a learnable parameter with an initial value of 0.5. The enhanced feature $\mathbf{F}_{\mathrm{enh},t}\in \mathbb{R}^{H'\times W'\times C}$ is fused with the original structural features $\mathbf{x}_t$, which is then fed into a feedforward network to get $\hat{M}_{raw,t} \in \mathbb{R}^{H'\times W'}$. After bilinear interpolation, we obtain a full-resolution probability map $\hat{M}_t \in \mathbb{R}^{H\times W} $. The magnitude of each pixel value in $\hat{M}_t$ directly indicates the probability of moving semantics.

\subsubsection{Training Progress}
\ 
\newline
\indent 
To train generalizable motion model capable of capturing spatiotemporal features, we design a supervised loss function leveraging ground truth $M_{gt,t}\in\mathbb{R}^{H\times W}$, defined as:
\begin{equation}
    \label{equ:distribution}
	\begin{aligned} 
\mathcal{L}_{\mathrm{GD}}=\lambda_1\mathcal{L}_{\mathrm{base}}+\lambda_2\mathcal{L}_{\mathrm{reg}}+\lambda_3\mathcal{L}_{\mathrm{dice}}
	\end{aligned}
\end{equation}
The base loss is computed as the pixel-wise absolute difference between $M_{gt,t}$ and $\hat{M}_t$. For ambiguous predictions in intermediate values (e.g., $0.5$), we introduce a binary entropy penalty $\mathcal{L}_{\mathrm{reg}}=-(\hat{M}_tlog(\hat{M}_t)+(1-\hat{M}_t)log(1-\hat{M}_t))$, which has maximum gradient magnitude at intermediate values, driving predictions to $\{0,1\}$ outputs. 

While $\mathcal{L}_{\mathrm{base}}$ employs pixel-wise optimization to maximize geometric fidelity, it exhibits limited capacity for structural feature preservation. To holistically capture shape integrity, we integrate the Dice loss function for regularization:
\begin{equation}
    \label{equ:distribution}
	\begin{aligned} 
\mathcal{L}_{\mathrm{dice}}=&1-\frac{2|M_{gt,t}\cap \hat{M}_t|}{|M_{gt,t}|+|\hat{M}_t|}\\
=&1-\frac{2\sum_{x,y}M_{gt,t}(x,y)\hat{M}_{t}(x,y)}{\sum_{x,y}(M_{gt,t}(x,y)+\hat{M}_{t}(x,y))}
	\end{aligned}
\end{equation}

Given that a probabilistic representation can introduce dynamically inaccurate data associations and that the ambiguous edges lead to poor tracking performance, we perform binarization followed by dilation to eliminate interference from both probability and edge indistinctness. Specifically, after training the generalizable motion model, we apply Otsu's adaptive thresholding during inference to binarize probability maps, yielding the raw mask $M_{raw,t}$. Subsequently, morphological dilation with a disk-shaped structuring element $\mathbf{K}_r$ refines the edges of dynamic objects, yielding a domain-agnostic prior binary mask $M_t$. 

\begin{algorithm}[t]
\caption{Generalizable Motion Model ($\boldsymbol{I}_t$)}
\label{alg:dyn-prior}
\LinesNumbered
\DontPrintSemicolon
\KwIn{Image $\boldsymbol{I}_t\in \mathbb{R}^{H\times W\times3}$} 
\KwOut{$M_t \in \mathbb{R}^{H \times W}$ }

\uIf{$\textbf{UnInit}()$}{
  $\mathcal{Q}_t \leftarrow [\mathbf{0}]^{\times L}$; 
\KwRet}

$\mathbf{x}_t  = \text{DINOv2}(\boldsymbol{I}_t)$ \tcp*{$\mathbf{x}_t\!\in\!\mathbb{R}^{H'\times W'\times C}$}

$\mathbf{Q}_t = \mathbf{x}_t \mathbf{W}^{Q}$;\quad
$\mathbf{K}_t = \mathcal{Q}_t \mathbf{W}^{K}$;\quad
$\mathbf{V}_t = \mathcal{Q}_t \mathbf{W}^{V}$

$\mathbf{F}_{\text{attn},t} = \text{MultiHead\_Attention}(\mathbf{Q}_t,\mathbf{K}_t,\mathbf{V}_t)$

$\mathbf{D} \leftarrow \text{Dyna\_Head}(\mathbf{F}_{\text{attn},t});$

$\mathbf{S} \leftarrow \text{Static\_Head}(\mathbf{F}_{\text{attn},t})$

$\mathbf{F}_{\mathrm{enh},t} = \mathbf{F}_{\text{attn},t} 
\odot \left( 1 + \alpha \mathbf{D} \right)
\odot \left( 1 - \alpha \mathbf{S} \right)$

$\hat{M}_{raw,t}=\text{Feed-Forward}(\mathbf{F}_{\mathrm{enh},t}\mathbf{W}^{enh}+\mathbf{x}_t)$
\;

$\hat{M}_t \leftarrow$ bilinear interpolation of $\hat{M}_{raw,t}$ to $(H,W)$

$\mathcal{Q}_{t}.\text{push}(\mathbf{x}_t)$

\uIf{$\textbf{Training}()$}{
  $\mathcal{L}_{\mathrm{GD}}=\lambda_1\mathcal{L}_{\mathrm{base}}+\lambda_2\mathcal{L}_{\mathrm{reg}}+\lambda_3\mathcal{L}_{\mathrm{dice}}$; 

  $Loss.backward()$; \KwRet
}
\ElseIf{$\textbf{Inference}()$}{
$\tau_{t}=\mathcal{O}_{\mathrm{otsu}}(\hat{M}_{t})$

$M_{raw,t}=\text{Binarize}(\hat{M}_{t},\tau_{t})$

$\mathbf{K}_r=\{(u,v)\mid u^2+v^2\le r^2\}$

$M_{t}(x,y)=\max_{(u,v)\in \mathbf{K}_r} M_{raw,t}(x-u,y-v)$\quad
}

\KwRet $M_t$ \tcp*{Final motion prior}
\end{algorithm}

\subsection{Tracking Progress}
\label{subsec:Location Algorithm}
Building upon established methodology, we incorporate scale-aware monocular depth estimates $\boldsymbol{D}_{est}\in\mathbb{R}^{H\times W}$  from Metric3D-v2 to facilitate robust pose estimation, leveraging its zero-shot generalization capability across diverse scenes.

For pose estimation, we adopt DROID-SLAM's dense bundle adjustment (DBA) monocular SLAM framework\cite{droid}, which operates over a frame graph $G=(V,\mathcal{E})$, where $V$ denotes selected keyframes and $\mathcal{E}$ represents their covisibility constraints. 
The goal is to optimize camera poses $\boldsymbol{T}_{[*]}$ and estimate monocular depth maps $d_{[*]}\in\mathbb{R}^{H\times W}$ for selected keyframes $V$. However, dynamic points induce incorrect factor graph construction, degrading computational efficiency and system performance. To mitigate this, we leverage our generalizable motion model to provide dynamic object priors, which entirely eliminate residual errors from dynamic regions. Specifically, we extract the static component $S_{[*]}=1-M_{[*]}$, and $\Sigma_{[*]}$ is the basic covariance weight derived from DROID-SLAM. This approach reformulates dynamic interference as tractable optimization constraints, enhancing both accuracy and robustness in dynamic SLAM scenarios, as formalized below.
\begin{equation}
    \label{equ:prior}
	\begin{aligned} 
\mathop{\arg\min}\limits_{\{\boldsymbol{T}_i,d_i\}_{i\in V}}&\Bigg\{\sum_{(i,j)\in\mathcal{E}}\left\|\mathbf{E}_{ij}(\boldsymbol{T}_i,\boldsymbol{T}_j,d_i,d_j)\right\|_{\Sigma_{ij}\odot S_i}^2+\\
\lambda_d\sum_{i\in V}&\left\|S_i\left(d_{i}-\boldsymbol{D}_{est,i} \right)\right\|^{2}+\lambda_{s}\sum_{t\in V}\left\|\log \left(\boldsymbol{T}_{t-1}^{-1}\boldsymbol{T}_{t}\right)\right\|^{2}\Bigg\}\\
	\end{aligned}
\end{equation}
where the first term represents the monocular pose estimation objective from DROID-SLAM, where residual weights in dynamic regions are proactively set to zero to enhance computational efficiency by excluding non-static contributions. The second term constitutes a depth supervision loss leveraging neural depth predictions, while the third term imposes trajectory smoothness regularization,  penalizing abrupt pose variations between consecutive frames.

\subsection{Mapping Progress}
\label{subsec:Mapping Algorithm}
\subsubsection{GMM Guided Uncertainty}
Based on the uncertainty-aware framework\cite{wildgs}, we process the features $\mathbf{x}_t$ through a shallow MLP $P$ to predict an uncertainty map $U_t=P(\mathbf{x}_t)\in \mathbb{R}^{H\times W}$. While this approach effectively handles ambiguous distractors and enhances rendering quality, its reliance on per-scene 3DGS rendering loss supervision often leads to dynamic misjudgment for single-frame inputs. To address this limitation, we integrate our generalizable motion model that aggregates historical features, embedding it as a temporal prior into the original framework as follows:
\begin{equation}
    \label{equ:distribution}
	\begin{aligned} 
\mathcal{L}_{\mathrm{uncer}}=\lambda_{1}\mathcal{L}_{\mathrm{3DGS}}+\lambda_{2} \mathcal{L}_{\mathrm{prior}}+\lambda_{3}\mathcal{L}_{\mathrm{reg\_U}}
	\end{aligned}
\end{equation}
where $\mathcal{L}_{\mathrm{3DGS}}$ quantifies the reconstruction error between rendered image and ground-truth inputs, as formally defined in (\ref{equ:3dgs}). The regularization term $\mathcal{L}_{\mathrm{reg\_U}}=\log U_t$ constrains uncertainty magnitudes, preventing them from approaching infinity. The error of our prior model is defined as:
\begin{equation}
    \label{equ:prior}
	\begin{aligned} 
\mathcal{L}_{\mathrm{prior}}= \sum_{x,y}M_{t}(x,y)\cdot\left|\tfrac{1}{U_t(x,y)}-T_{\mathrm{max}}\right|
	\end{aligned}
\end{equation}
where $T_{\mathrm{max}}$ denotes the target uncertainty threshold for dynamic regions. Equation (\ref{equ:prior}) effectively mitigates inaccuracies in uncertainty-aware dynamic object identification while preserving non-disruptive processing of static distractors like noisy observations or varying light conditions.

\subsubsection{Incremental Gaussian Map}
\ 
\newline
\indent 
After obtaining a new keyframe, we incrementally create Gaussians to optimize the map. For newly observed feature points in the image, we add a new Gaussian with the color $\boldsymbol{c}_*$ of that pixel, the location $\boldsymbol{\mu}_*$ obtained by unprojecting the pixel, an opacity $\alpha_*$ of 0.5, and a radius of 0.1. 

When newly observed frames contain dynamic objects, we perform stochastic neighborhood sampling within these regions to preserve geometric continuity in occluded areas. Specifically, we construct a KD-tree for the $(\boldsymbol{\mu}_{*,x} , \boldsymbol{\mu}_{*,y})$ of static Gaussians from image $\boldsymbol{I}_t$. For dynamic point $\boldsymbol{\mu}_i\in M_t$, we query $(\boldsymbol{\mu}_{i,x} , \boldsymbol{\mu}_{i,y})$ with its k-nearest neighbors and perform random sampling within this local neighborhood to replace the depth and color attributes by those of static points, mitigating occlusion artifacts as follows:
\begin{equation}
    \label{equ2:3dgs}
	\begin{aligned}
\boldsymbol{\mu}_{i}\leftarrow(\boldsymbol{\mu}_{i,x} ,\boldsymbol{\mu}_{i,y} ,\boldsymbol{\mu}_{j,z})\;\; \boldsymbol{c}_{i}\leftarrow \boldsymbol{c}_{j},\;\;(j\sim\text{Sample}(1,k))
        \end{aligned}
\end{equation}
We then apply scale expansion and opacity enhancement to occluded points, mitigating optimization efficiency deterioration caused by point cloud sparsification in occluded regions.
\begin{figure}[t]
    \centering
    \includegraphics[width=8.5cm]{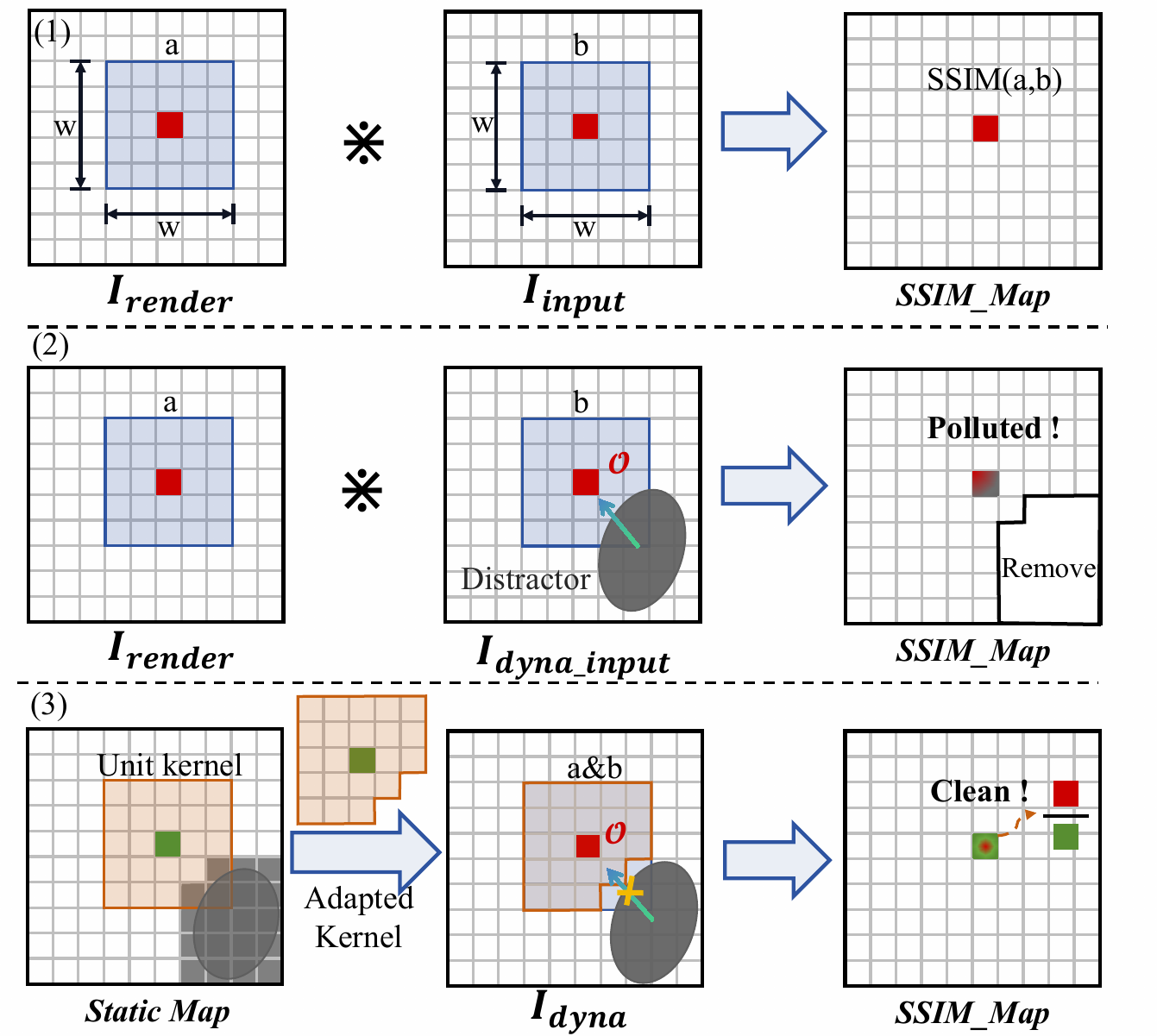}
    \caption{Illustration of distractor-adaptive
SSIM:(1) Computation of SSIM; (2) Conventional dynamic SSIM processing is polluted; (3) Our adapted kernel achieves clean SSIM under distractors.}
    \label{SSIM}
    \vspace{-0.2cm}
\end{figure}
\subsubsection{Gaussian Update}
\ 
\newline
\indent 
Gaussian map renders RGB images by sorting 3D Gaussians based on their view depth, followed by $\alpha$-blending 2D projections of the $\mathcal{N}$ Gaussians. For each pixel, the color $\boldsymbol{I}_{r}$ and depth $\boldsymbol{D}_{r}$ are determined through this blending process:
\begin{equation}
  \begin{aligned}
    \boldsymbol{I}_{r}=\sum_{i\in \mathcal{N}}\mathbf{c}_i\alpha_i\prod_{j=1}^{i-1}(1-\alpha_j), \boldsymbol{D}_{r}=\sum_{i\in \mathcal{N}}\mathbf{d}_i\alpha_i\prod_{j=1}^{i-1}(1-\alpha_j)
  \end{aligned}
\end{equation}

The Gaussian parameters are then updated iteratively by gradient-based optimization to minimize the mapping loss:
\begin{equation}
    \label{equ:motion model}
	\begin{aligned}
\mathcal{L}_{\mathrm{mapping}}=\lambda_{1}\mathcal{L}_{\mathrm{3DGS}}+\lambda_2\mathcal{L}_{\mathrm{iso}}+\lambda_3\mathcal{L}_{\mathrm{ssim\_dy}}
        \end{aligned}
\end{equation}

$\mathcal{L}_{\mathrm{3DGS}}$ denotes the residual metrics between the 3DGS-rendered output and inputs, which are constrained by an uncertainty map through element-wise division as follows. And we employ $\mathcal{L}_{\mathrm{iso}}$ regularization\cite{gsslam} on the scaling parameters to suppress artefacts in sparsely observed regions.
\begin{equation}
    \label{equ:3dgs}
	\begin{aligned}
\mathcal{L}_{\mathrm{3DGS}}=(1-\lambda_{\mathrm{d}})\left|\frac{\boldsymbol{I}_{r}-\boldsymbol{I}}{U_t^2}\right|_1+\lambda_{\mathrm{d}}\left|\frac{\boldsymbol{D}_r-\boldsymbol{D}_{est}}{U_t^2}\right|_1
        \end{aligned}
\end{equation}

In contrast to pixel-wise difference loss, $\mathcal{L}_{\mathrm{ssim}}$—as illustrated in Fig. \ref{SSIM}(1)—preserves structural coherence across local regions by jointly modeling luminance (mean $\mu$), contrast (deviation $\sigma_{i, i}$), and spatial relationships (covariance $\sigma_{i,j}$). In dynamic scenes, although traditional methods remove SSIM values from distractor regions after computing the SSIM map, the red pixel $\mathcal{O}$ in Fig. \ref{SSIM}(2) is not within the dynamic range and thus not removed, leading to an inaccurate SSIM loss being introduced at the red pixel $\mathcal{O}$. To address this issue, we propose a distractor-adaptive SSIM loss designed for dynamic scenes as Fig. \ref{SSIM}(3). Specifically, we perform a Hadamard product and convolution on the prior static component $S_t$ using a unit kernel $w_\mathrm{unit}$, which can be converted into an adapted kernel $w_{\mathrm{ad}}(\mathcal{O})$ while obtaining the number of valid static pixels $N_{\mathrm{ad}}(\mathcal{O})$. By using the generated kernel for SSIM computation and normalizing valid pixels, we avoid the introduction of dynamic loss, thereby yielding an accurate SSIM loss specific to pure static regions. The calculation formula for the luminance ($\mu$) of red pixel $\mathcal{O}$ is as follows: 
\begin{subequations}
  \begin{align}
    \mathrm{SSIM\_Map}&_{\mu}(\mathcal{O})=\frac{(w_{\mathrm{ad}}(\mathcal{O})\odot  w_{\mathrm{ori},\mu})*\boldsymbol{I}}{N_{\mathrm{ad}}(\mathcal{O})}\\
     w_{\mathrm{ad}}(\mathcal{O})=w_{\mathrm{unit}}&\odot  S_t(\mathcal{O}),N_{\mathrm{ad}}(\mathcal{O})=w_{\mathrm{unit}}* S_t(\mathcal{O})
  \end{align}
\end{subequations}
where $S_t(\mathcal{O})$ denotes the local region of $S_t$ that matches the kernel $w_{\mathrm{unit}}$ size at pixel $\mathcal{O}$. The calculation of both contrast and spatial relationships is replaced using the generated kernel $w_\mathrm{ad}(\mathcal{O})$—in the same manner as the luminance($\mu$)— ultimately obtaining the distractor-adaptive SSIM map.

\begin{figure*}[t]
    \centering
\includegraphics[width=17cm]{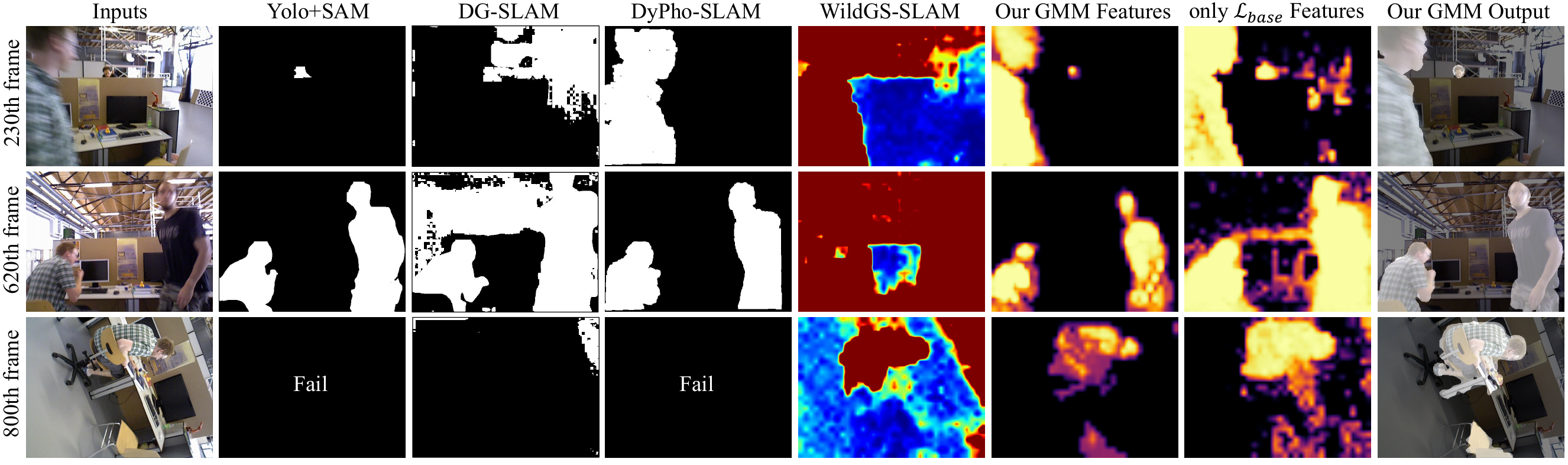}
    \caption{Qualitative results of the different dynamic extractors progressing in fr3/w/half.}
    \label{mask}
    \vspace{-0.2cm}
\end{figure*}

\begin{table*}[t]
\caption{\textbf{Camera tracking results on dynamic scenes from the TUM and Bonn challenging datasets. }The units for ATE$(\downarrow)$ and Std.$(\downarrow)$ are in cm. The best results among \textit{Mono Dense SLAM} domains are in \textbf{bold}. $-$ indicates that the relevant algorithms are not yet open-source, and the corresponding data is not provided in the original report.}
\centering
\resizebox{0.99\textwidth}{!}{
        \large
        \begin{tabular}{c|cc|cc| cc|cc|cc| cc| cc|cc| cc}
\toprule
\multirow{2}{*}{\textbf{Method}}&\multicolumn{8}{c|}{\textbf{TUM}}&\multicolumn{10}{c}{\textbf{Bonn}}\\

&\multicolumn{2}{c}{\textbf{fr3/w/xyz}}&\multicolumn{2}{c}{\textbf{fr3/w/half}}&\multicolumn{2}{c}{\textbf{fr3/s/half}}&\multicolumn{2}{c|}{\textbf{Avg.}}&\multicolumn{2}{c}{\textbf{balloon}}&\multicolumn{2}{c}{\textbf{balloon2}}&\multicolumn{2}{c}{\textbf{ps\_track}}&\multicolumn{2}{c}{\textbf{crowd2}}&\multicolumn{2}{c}{\textbf{Avg.}}\\
\midrule
\textit{RGBD Dense SLAM} & ATE & Std.& ATE & Std.& ATE & Std. & ATE & Std. & ATE & Std. & ATE & Std. & ATE & Std.& ATE & Std. & ATE & Std. \\
RoDyn-SLAM\cite{rodyn-slam}&8.3 & 5.5 & 5.6 & 2.8 & 4.4 & 2.2 & 4.1 & 2.3 & 7.9 & 2.7 & 11.5 & 6.1 & 14.5 & 4.6 & 6.5 & 3.9 & 12.3 & 4.4\\
GARAD-SLAM\cite{GARAD}&1.6 &0.8 & 2.3 & 1.2 & - & - & 1.9 & 1.1 & 3.0 & 1.2 & 2.5 & 1.2 & 4.6 & 1.5 & 2.4 & 1.2 & 2.7 & 1.2\\
Gassidy\cite{Gassidy}&3.5 & 1.6 & 3.7 & 1.9 & 2.4 & 1.4 & 2.6 & 1.3 & 2.6& 0.8 & 7.6 & 3.4 & 10.3 & 4.4 & - & - & 7.8 & 3.1\\
DGS-SLAM\cite{dgs-slam}&4.1 & 2.2 & 5.5 & 2.8 & 4.1 & 1.6 & 3.0 & 1.5 & 2.9 & 0.8 & 6.0 & 2.8 & 9.8 & 4.1 & - & - & 7.3 & 3.0\\
ADD-SLAM\cite{addslam}&1.4 & 0.9 & 1.6 & 0.8 & 1.3 & 0.6 & 1.3 & 0.7 & 2.7 & 1.1 & 2.3 & 0.8 & 3.7 & 1.1 & - & - & 2.8 & 1.1\\
UP-SLAM\cite{upslam}&1.6 &-& 2.6 &-&-&-& 1.4 &-& 2.8 &-& 2.7 &-& 4.0&-&-&-& 3.2&-\\
DG-SLAM\cite{dgslam}&1.8 & 1.1 & 2.0& 1.0& 2.3 & 1.6 & 2.2 & 1.0& 3.7 & 1.6 & 4.1 & 1.7 & 4.3 & 1.9 & 6.8 & 4.2 & 5.5 & 3.9\\
DyPho-SLAM\cite{dypho-slam}&1.6 & 0.8 & 2.6 & 1.3 & 1.6 & 0.7 & 1.6 & 0.7 & 3.0 & 1.2 & 2.7 & 1.3 & 3.7 & 1.4 & 2.5 & 1.3 & 3.2 & 1.5\\

\midrule
\textit{Mono Dense SLAM}  &ATE&Std.&ATE&Std.&ATE&Std.&ATE&Std.&ATE&Std.&ATE&Std.&ATE&Std.&ATE&Std.&ATE&Std.\\
Splatam\cite{splatam}&136.6 & 33.7 & 185.4 & 67.6 & 14.1 & 6.8 & 102.6 & 40.8 & 35.7 & 14.3 & 36.8 & 16.1 & 158.2 & 55.3 & 35.8 & 19.6 & 63.7 & 28.8\\
MonoGS\cite{gsslam}&40.9 & 16.3 & 11.3 & 7.2 & 8.2 & 3.9 & 24.9 & 14.0 & 35.4 & 17.4 & 28.9 & 15.8 & 26.0 & 14.6 & 7.4 & 3.0 & 27.5 & 15.1\\
DynaMoN(MS\&SS)\cite{dynamon} &1.4 &-& 1.9 &-& 2.3 &-& 1.6 &-& 2.8 &-& 2.7 &-& 14.8 &-& 2.8 &-& 5.2&-\\
Method in \cite{Sensor} &1.2 &-& 2.3 &-& 2.2 &-& 2.4 &-& - &-& - &-& \textbf{1.2} &-& 2.9 &-& 3.7&-\\
Dy3DGS-SLAM\cite{dy3dgs}&5.8 &-& 7.0 &-& 3.4 &-& 4.7 &-& 4.5 &-& \textbf{1.9} &-& 5.6 &-& - &-& 4.5&-\\
WildGS-SLAM\cite{wildgs} &1.3 & \textbf{0.6} & 1.5 & 0.8 & 1.8 & 1.1 & 1.4 & 0.8 & 2.9 & 1.2 & 2.5 & 1.2 & 3.6 & 1.9 & 2.3 & 1.1 & 2.9 & 1.4 \\
Ours& \textbf{1.1} & \textbf{0.6} & \textbf{1.4} & \textbf{0.7} & \textbf{1.5} & \textbf{0.8} & \textbf{1.3} & \textbf{0.7} & \textbf{2.4} & \textbf{1.0} & 2.3 & \textbf{1.1} & 3.4 & \textbf{1.8} & \textbf{1.8} & \textbf{0.8} & \textbf{2.7} & \textbf{1.1}\\
\bottomrule
\end{tabular}
}

    \vspace{-0.2cm}
\label{track}
\end{table*}

\section{EXPERIMENTS}
\subsection{Experiment Setup }

\textbf{Implementation details}: In our experiments, we adopt WildGS-SLAM\cite{wildgs} as the baseline, executing both GGD-SLAM and the training of the generalizable motion model on a server equipped with a 3.00GHz Intel Xeon Gold 6151 CPU and consumer-grade GPU. We also locally rerun classic monocular 3DGS-SLAM algorithms: MonoGS\cite{gsslam} and Splatam\cite{splatam}, and reimplement SOTA RGBD-based 3DGS-SLAM frameworks adapted to dynamic environments: DG-SLAM\cite{dgslam} and DyPho-SLAM\cite{dypho-slam}. The key parameters are set as follows: $L=12,H'=384,W'=512,k=10,\lambda_{\mathrm{d}}=0.5, T_{max}=0.1$.

\textbf{Datasets}: We evaluate our approach using three prominent dynamic SLAM datasets: the TUM RGB-D Dataset \cite{tum}, the Bonn RGB-D
Dynamic Dataset \cite{bonn}, and the Wild-SLAM Dataset\cite{wildgs}, which are captured in real-world environments with a handheld device, providing RGB images and ground truth trajectories. We use the Davis Dataset\cite{davis} to train our generalizable motion model, which contains ground-truth motion masks.

\textbf{Metric}: We evaluate camera tracking performance using the Root Mean Square Error (RMSE) and Standard Deviation (Std.) of Absolute Trajectory Error (ATE) \cite{tum}. To evaluate the reconstruction quality, we first perform a qualitative assessment using novel view renderings and then quantify it by measuring the average difference between the rendered and input images in static regions only.

\subsection{Evaluation of Tracking}
\label{subsec: Real-world experiments}
To demonstrate our generalizable dynamic semantic extraction network, we present a visualization example for the fr3/w/half sequence as Fig. \ref{mask}. Specifically, misclassification is prone to occur when using single-image specific label segmentation for small target recognition, blurry fast-moving objects, and large-scale camera movements. The uncertainty-awareness of WildGS-SLAM is highly susceptible to misjudgment in background regions, leading to insufficient effective image information and thus impairing both localization output and background reconstruction. In contrast, our generalizable dynamic semantic extraction network can extract moving object semantics based on the historical frames, achieving excellent extraction performance. We visualize the GMM feature $\hat{M}_{t}$ as ``Our GMM Features", while using only the $\mathcal{L}_{\mathrm{base}}$ for pixel-wise learning limits structural feature preservation, resulting in significant noise. In the 800th frame, while the chair is static in a single frame, it moves over long-term historical observations—so the model accurately classifies it as dynamic in the current scene. 

\begin{table}[t]
    \caption{\textbf{Ablation study on Bonn RGB-D
Dynamic Dataset. }Generalizable Prior refers to prior information from GMM; OTSU Binarize is our solution for ambiguous edges; and Smoothness refers to the smoothness term in Tracking.
  }  
\centering
\resizebox{0.46\textwidth}{!}{
        \large
        
     \begin{tabular}{c|c|c|cc}  
    \toprule
    Generalizable Prior&OTSU Binarize& Smoothness & ps\_track & crowd2  \\  
    \midrule
    \checkmark &\checkmark & $\times$ & 3.47 & 1.95  \\ 
    \checkmark &$\times$ & \checkmark & 3.44 & 1.86  \\  
    $\times$ & $\times$ &\checkmark & 3.56 & 2.14  \\  
    \checkmark & \checkmark & \checkmark &\textbf{3.41} & \textbf{1.79} \\  
    \bottomrule
   
  \end{tabular}

}
    \vspace{-1em}
\label{Co}
\end{table}

We apply the prior information obtained from the generalizable motion model to the tracking process, and compare our method with state-of-the-art dynamic dense SLAM approaches in Table \ref{track}. RGB-D-based methods show competitive performance in localization precision, primarily due to the precise absolute scale information provided by the depth sensor. Although specialized dynamic Mono-based SLAM systems such as Dy3DGS-SLAM and WildGS-SLAM incorporate the dynamic object handling capabilities, they still show inferior performance compared with our method. This is primarily because they lack a precise dynamic distractor recognition method and introduce incorrect data association. Our method combines an efficient generalizable motion model for progressive SLAM inputs, enabling exceptional tracking precision even better than the RGB-D-based method in highly dynamic sequences such as “fr3/w/half” and “bonn/crowd2”. Additionally, an ablation study is conducted on the TUM and Bonn datasets to evaluate the proposed method, with the ATE results in Table \ref{Co} confirming the robustness of the related module.
\begin{figure*}[t]
    \centering
    \includegraphics[width=18cm]{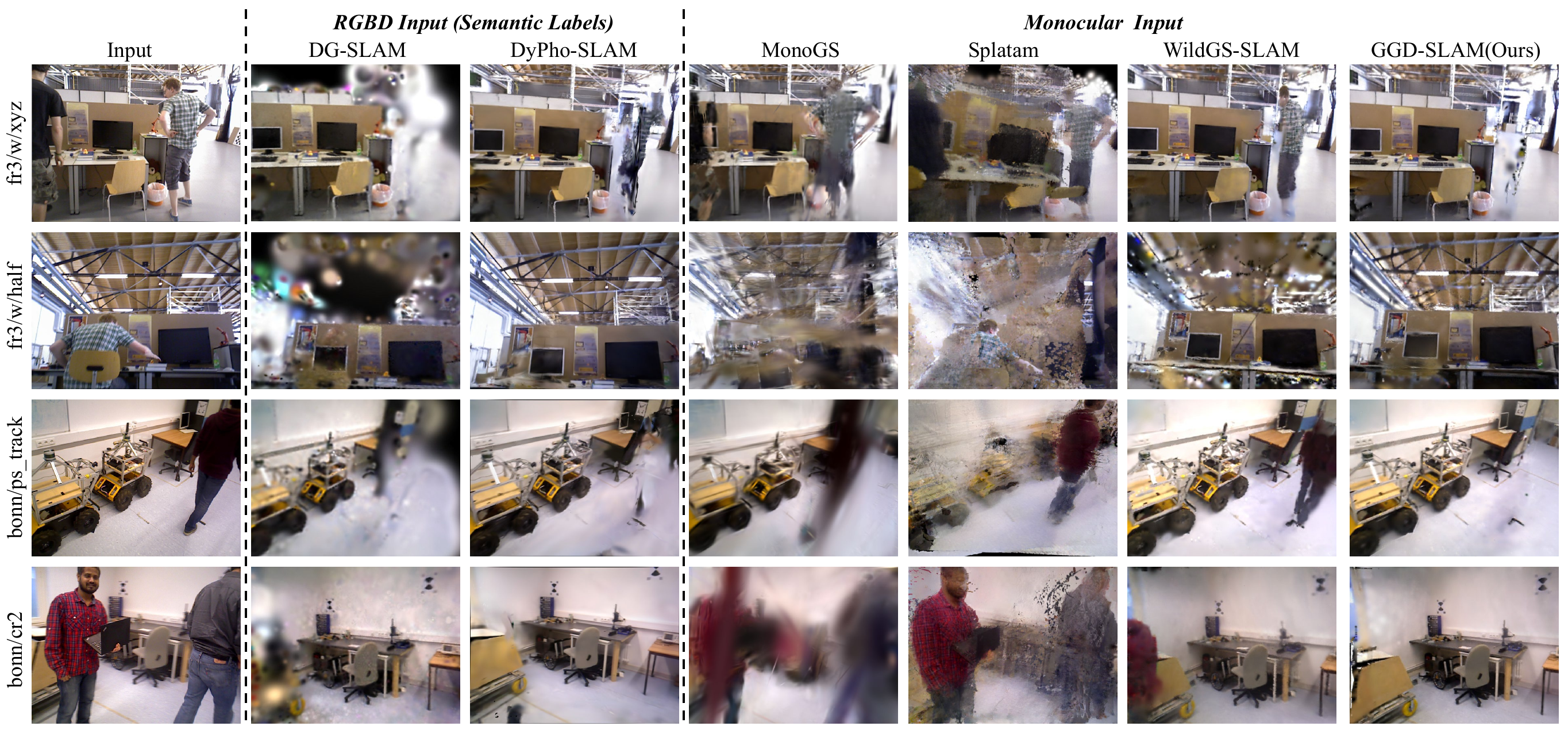}
    \caption{ Comparison of rendered results from state-of-the-art Gaussian Splatting SLAM methods. }
    \label{mapping}
    \vspace{-0.2cm}
\end{figure*}
\begin{table*}[t]
\centering
\caption{\textbf{Mapping results on dynamic scenes from the TUM and Bonn dynamic datasets.} The unit for PSNR($\uparrow$) is in dB. SSIM($\uparrow$) and LPIPS($\downarrow$)  are both dimensionless metrics.}
\resizebox{0.9\textwidth}{!}{
        \large
        
\begin{tabular}{c|ccc|ccc| ccc| ccc| ccc}
\toprule
\multirow{2}{*}{\textbf{Method}}&\multicolumn{3}{c|}{\textbf{fr3/w/xyz}}&\multicolumn{3}{c|}{\textbf{fr3/w/half}}&\multicolumn{3}{c|}{\textbf{bonn/ps\_tracking}}&\multicolumn{3}{c|}{\textbf{bonn/crowd2}}&\multicolumn{3}{c}{\textbf{Avg.}}\\
&PSNR&SSIM&LPIPS&PSNR&SSIM&LPIPS&PSNR&SSIM&LPIPS&PSNR&SSIM&LPIPS&PSNR&SSIM&LPIPS\\
\midrule
MonoGS & 13.97 & 0.467 & 0.591 & 16.57 & 0.551 & 0.476 & 19.56 & 0.726 & 0.454 & 18.07 & 0.695 & 0.567 & 17.04 & 0.610 & 0.522 \\
Splatam & 14.29 & 0.527 & 0.486 & 15.53 & 0.478 & 0.493 & 17.81 & 0.596 & 0.344 & 15.79 & 0.517& 0.592 & 15.81 & 0.526 & 0.479\\
WildGS-SLAM &21.53 & \textbf{0.821} & 0.141 & 20.89 & 0.829 & 0.136 & \textbf{23.35} & \textbf{0.904} & \textbf{0.148} & 23.15 & 0.885 & 0.264 & 22.23 & 0.855 & 0.172 \\
Ours& \textbf{22.68} & 0.751 & \textbf{0.117} & \textbf{22.25} & \textbf{0.848} & \textbf{0.134} & 22.92 & 0.902 & 0.162 & \textbf{24.27} & \textbf{0.918} & \textbf{0.219} & \textbf{23.03} & \textbf{0.859} & \textbf{0.158}  \\

\bottomrule
\end{tabular}
}

    \vspace{-0.2cm}
\label{allpsnr}
\end{table*}

 \subsection{Evaluation of Mapping}
\label{subsec: Real-world experiments}
To assess the mapping capabilities, we benchmark our method against open-source 3DGS SLAM algorithms. As shown in Fig. \ref{mapping}, DG-SLAM and DyPho-SLAM require semantic labels and depth input. MonoGS and Splatam suffer severe performance degradation under dynamic interference. WildGS-SLAM depends on 3DGS rendering, which results in poor background rendering after large-scale camera movements (e.g., fr3/w/half), thereby leading to misjudgment of dynamic objects and subsequent system deterioration. Additionally, in other scenarios, it exhibits inadequate removal of occlusions for edge objects, causing the occurrence of residual artifacts. In contrast, our proposed GGD-SLAM can effectively eliminate dynamic interference while maintaining high-quality background rendering.  As shown in Table \ref{allpsnr}, our proposed SLAM achieves the
 best performance among monocular GS-based frameworks on the dynamic sequences of TUM and Bonn challenging datasets. In addition, we conduct ablation experiments for our proposed method. The specific PSNR results, as shown in Table \ref{ablition}, demonstrate the effectiveness of our mapping module.
 
\begin{table}[t]  
  \centering  
   \caption{\textbf{Ablation study on distractor-adaptive SSIM and static Gaussian KD-tree for our occlusion recovery method.}
  }  
  \begin{tabular}{c|c|cc}  
    \toprule
    Dynamic SSIM& Static KD-Tree & fr3/w/xyz & bonn/cr2  \\  
    \midrule
    \checkmark & $\times$ & 21.96 & 23.75  \\  
    $\times$ & \checkmark & 21.59 & 23.47  \\  
    \checkmark & \checkmark & \textbf{22.68} & \textbf{24.27} \\  
    \bottomrule
   
  \end{tabular}
 
  \label{ablition}  
   \vspace{-0.2cm}
\end{table}

\subsection{More Generalized Scenarios}
\begin{figure}[t]
    \centering
    \includegraphics[width=8.5cm]{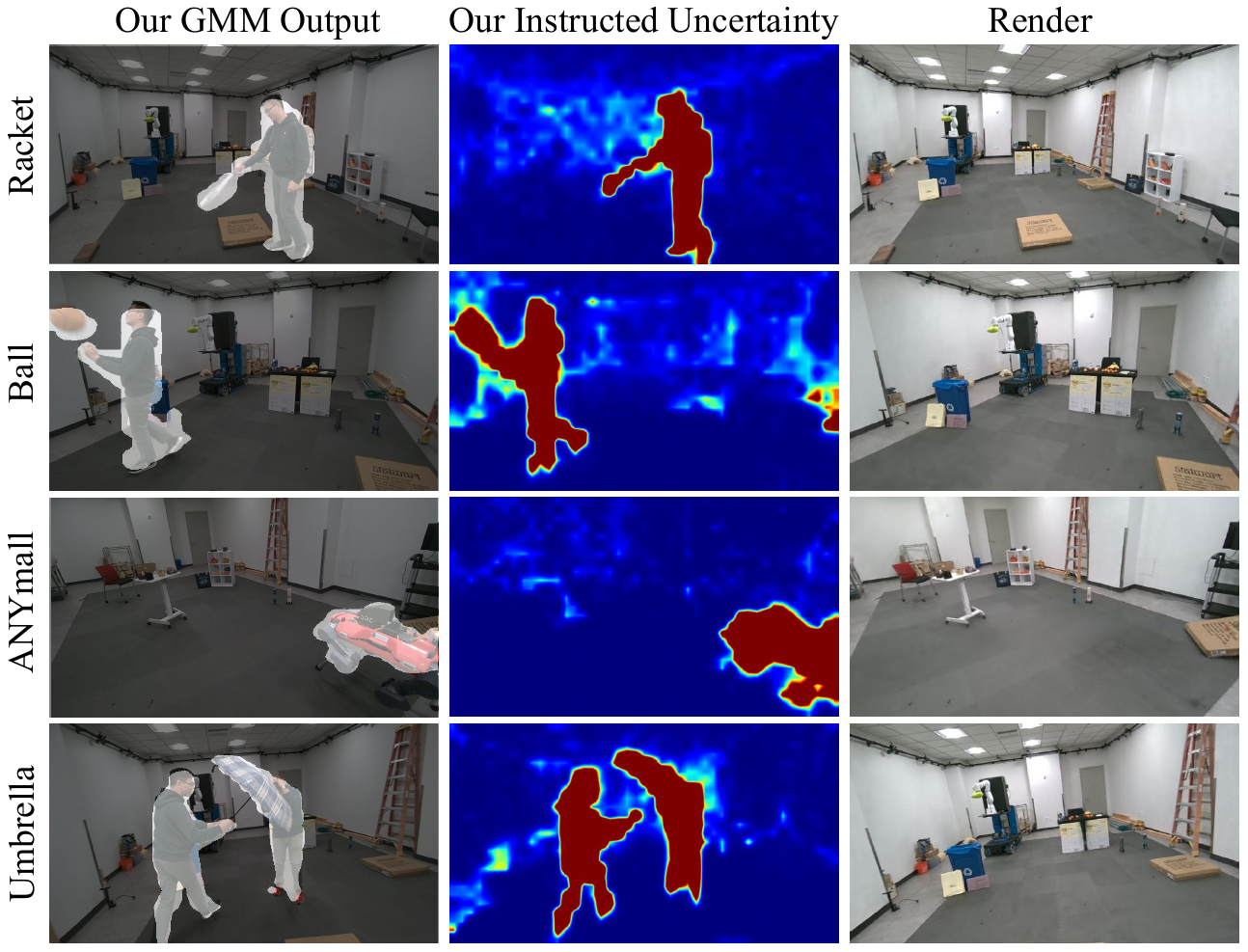}
    \caption{Our GGD-SLAM performance on the Wild-SLAM Dataset.}
    \label{wildgs}
    \vspace{-0.2cm}
\end{figure}

As shown in Fig. \ref{wildgs}, we evaluate our generalizable dynamic semantic extraction network on the more generalized Wild-SLAM Dataset, where it successfully segments moving objects of various types and guides the uncertainty generation to achieve high-quality rendering. Compared to the TUM and Bonn datasets, the Wild-SLAM Dataset offers higher-resolution imagery and milder camera motion, making it easy to achieve superior 3DGS rendering quality. Under these conditions, the uncertainty-aware mechanism adapts effectively, resulting in both GGD-SLAM and WildGS-SLAM delivering exceptionally high performance.

\section{Conclusion and Future Work}
\label{sec:conclusion}
This study proposes GGD-SLAM, a generalizable and robust framework to achieve localization and photorealistic mapping in dynamic environments. A generalizable dynamic extractor method, which leverages attention mechanisms on a queue of historical frames to extract dynamic semantics, is proposed to address the challenge of dynamic removal. Furthermore, it is integrated with a background consistency mapping process to minimize the impact of dynamic objects on static components. Extensive experiments have demonstrated that GGD-SLAM significantly outperforms existing SOTA SLAMs for photorealistic mapping. In the future, we aim to develop a method for real-time reconstruction of dynamic object motion and inpainting of fully occluded regions, while ensuring the stability of the static scene.

{
    \balance
    \bibliographystyle{IEEEtran}
    \bibliography{IEEEabrv, main}

\begin{thebibliography}{10}
\providecommand{\url}[1]{#1}
\csname url@samestyle\endcsname
\providecommand{\newblock}{\relax}
\providecommand{\bibinfo}[2]{#2}
\providecommand{\BIBentrySTDinterwordspacing}{\spaceskip=0pt\relax}
\providecommand{\BIBentryALTinterwordstretchfactor}{4}
\providecommand{\BIBentryALTinterwordspacing}{\spaceskip=\fontdimen2\font plus
\BIBentryALTinterwordstretchfactor\fontdimen3\font minus \fontdimen4\font\relax}
\providecommand{\BIBforeignlanguage}[2]{{%
\expandafter\ifx\csname l@#1\endcsname\relax
\typeout{** WARNING: IEEEtran.bst: No hyphenation pattern has been}%
\typeout{** loaded for the language `#1'. Using the pattern for}%
\typeout{** the default language instead.}%
\else
\language=\csname l@#1\endcsname
\fi
#2}}
\providecommand{\BIBdecl}{\relax}
\BIBdecl

\bibitem{2}
J.~Cheng, L.~Zhang, Q.~Chen, X.~Hu, and J.~Cai, ``A review of visual slam methods for autonomous driving vehicles,'' \emph{Engineering Applications of Artificial Intelligence}, vol. 114, p. 104992, 2022.

\bibitem{3}
F.~Tosi, Y.~Zhang, Z.~Gong, E.~Sandstr{\"o}m, S.~Mattoccia, M.~R. Oswald, and M.~Poggi, ``How nerfs and 3d gaussian splatting are reshaping slam: a survey,'' \emph{arXiv preprint arXiv:2402.13255}, vol.~4, 2024.

\bibitem{Gassidy}
L.~Wen, S.~Li, Y.~Zhang, Y.~Huang, J.~Lin, F.~Pan, Z.~Bing, and A.~Knoll, ``Gassidy: Gaussian splatting slam in dynamic environments,'' \emph{arXiv preprint arXiv:2411.15476}, 2024.

\bibitem{addslam}
W.~Wu, C.~Su, S.~Zhu, T.~Deng, Z.~Liu, and H.~Wang, ``Add-slam: Adaptive dynamic dense slam with gaussian splatting,'' \emph{arXiv preprint arXiv:2505.19420}, 2025.

\bibitem{dypho-slam}
Y.~Liu, K.~Fan, B.~Lan, and H.~Liu, ``Dypho-slam: Real-time photorealistic slam in dynamic environments,'' in \emph{2025 IEEE International Conference on Multimedia and Expo (ICME)}.\hskip 1em plus 0.5em minus 0.4em\relax IEEE, 2025, pp. 1--6.

\bibitem{wildgs}
J.~Zheng, Z.~Zhu, V.~Bieri, M.~Pollefeys, S.~Peng, and I.~Armeni, ``Wildgs-slam: Monocular gaussian splatting slam in dynamic environments,'' in \emph{Proceedings of the Computer Vision and Pattern Recognition Conference}, 2025, pp. 11\,461--11\,471.

\bibitem{sam}
A.~Kirillov, E.~Mintun, N.~Ravi, H.~Mao, C.~Rolland, L.~Gustafson, T.~Xiao, S.~Whitehead, A.~C. Berg, W.-Y. Lo \emph{et~al.}, ``Segment anything,'' in \emph{Proceedings of the IEEE/CVF international conference on computer vision}, 2023, pp. 4015--4026.

\bibitem{orbslam}
C.~Campos, R.~Elvira, J.~J.~G. Rodr{\'\i}guez, J.~M. Montiel, and J.~D. Tard{\'o}s, ``Orb-slam3: An accurate open-source library for visual, visual--inertial, and multimap slam,'' \emph{IEEE Transactions on Robotics}, vol.~37, no.~6, pp. 1874--1890, 2021.

\bibitem{dynaslam}
B.~Bescos, J.~M. F{\'a}cil, J.~Civera, and J.~Neira, ``Dynaslam: Tracking, mapping, and inpainting in dynamic scenes,'' \emph{IEEE Robotics and Automation Letters}, vol.~3, no.~4, pp. 4076--4083, 2018.

\bibitem{rodyn-slam}
H.~Jiang, Y.~Xu, K.~Li, J.~Feng, and L.~Zhang, ``Rodyn-slam: Robust dynamic dense rgb-d slam with neural radiance fields,'' \emph{IEEE Robotics and Automation Letters}, 2024.

\bibitem{5}
B.~Kerbl, G.~Kopanas, T.~Leimk{\"u}hler, and G.~Drettakis, ``3d gaussian splatting for real-time radiance field rendering.'' \emph{ACM Trans. Graph.}, vol.~42, no.~4, pp. 139--1, 2023.

\bibitem{10}
S.~Zhu, G.~Wang, D.~Kong, and H.~Wang, ``3d gaussian splatting in robotics: A survey,'' \emph{arXiv preprint arXiv:2410.12262}, 2024.

\bibitem{gsslam}
H.~Matsuki, R.~Murai, P.~H. Kelly, and A.~J. Davison, ``Gaussian splatting slam,'' in \emph{Proceedings of the IEEE/CVF Conference on Computer Vision and Pattern Recognition}, 2024, pp. 18\,039--18\,048.

\bibitem{splatam}
N.~Keetha, J.~Karhade, K.~M. Jatavallabhula, G.~Yang, S.~Scherer, D.~Ramanan, and J.~Luiten, ``Splatam: Splat track \& map 3d gaussians for dense rgb-d slam,'' in \emph{Proceedings of the IEEE/CVF Conference on Computer Vision and Pattern Recognition}, 2024, pp. 21\,357--21\,366.

\bibitem{dgslam}
Y.~Xu, H.~Jiang, Z.~Xiao, J.~Feng, and L.~Zhang, ``Dg-slam: Robust dynamic gaussian splatting slam with hybrid pose optimization,'' \emph{arXiv preprint arXiv:2411.08373}, 2024.

\bibitem{upslam}
W.~Zheng, L.~Ou, J.~He, L.~Zhou, X.~Yu, and Y.~Wei, ``Up-slam: Adaptively structured gaussian slam with uncertainty prediction in dynamic environments,'' \emph{arXiv preprint arXiv:2505.22335}, 2025.

\bibitem{dgs-slam}
M.~Kong, J.~Lee, S.~Lee, and E.~Kim, ``Dgs-slam: Gaussian splatting slam in dynamic environment,'' \emph{arXiv preprint arXiv:2411.10722}, 2024.

\bibitem{mos}
B.~Hou, Y.~Liu, N.~Ling, Y.~Ren, and L.~Liu, ``A survey of efficient deep learning models for moving object segmentation,'' \emph{APSIPA Transactions on Signal and Information Processing}, vol.~12, no.~1, 2023.

\bibitem{megasam}
Z.~Li, R.~Tucker, F.~Cole, Q.~Wang, L.~Jin, V.~Ye, A.~Kanazawa, A.~Holynski, and N.~Snavely, ``Megasam: Accurate, fast and robust structure and motion from casual dynamic videos,'' in \emph{Proceedings of the Computer Vision and Pattern Recognition Conference}, 2025, pp. 10\,486--10\,496.

\bibitem{monst3r}
J.~Zhang, C.~Herrmann, J.~Hur, V.~Jampani, T.~Darrell, F.~Cole, D.~Sun, and M.-H. Yang, ``Monst3r: A simple approach for estimating geometry in the presence of motion,'' \emph{arXiv preprint arXiv:2410.03825}, 2024.

\bibitem{segment}
N.~Huang, W.~Zheng, C.~Xu, K.~Keutzer, S.~Zhang, A.~Kanazawa, and Q.~Wang, ``Segment any motion in videos,'' in \emph{Proceedings of the Computer Vision and Pattern Recognition Conference}, 2025, pp. 3406--3416.

\bibitem{droid}
Z.~Teed and J.~Deng, ``Droid-slam: Deep visual slam for monocular, stereo, and rgb-d cameras,'' \emph{Advances in neural information processing systems}, vol.~34, pp. 16\,558--16\,569, 2021.

\bibitem{GARAD}
M.~Li, W.~Chen, N.~Cheng, J.~Xu, D.~Li, and H.~Wang, ``Garad-slam: 3d gaussian splatting for real-time anti dynamic slam,'' \emph{arXiv preprint arXiv:2502.03228}, 2025.

\bibitem{dynamon}
N.~Schischka, H.~Schieber, M.~A. Karaoglu, M.~Gorgulu, F.~Gr{\"o}tzner, A.~Ladikos, N.~Navab, D.~Roth, and B.~Busam, ``Dynamon: Motion-aware fast and robust camera localization for dynamic neural radiance fields,'' \emph{IEEE Robotics and Automation Letters}, 2024.

\bibitem{Sensor}
H.~Zhou, J.~Chen, and Z.~Li, ``Dynamic slam with 3d gaussian splatting supporting monocular sensing,'' \emph{IEEE Sensors Journal}, 2025.

\bibitem{dy3dgs}
M.~Li, Y.~Zhou, H.~Zhou, X.~Hu, F.~Roemer, H.~Wang, and A.~Osman, ``Dy3dgs-slam: Monocular 3d gaussian splatting slam for dynamic environments,'' \emph{arXiv preprint arXiv:2506.05965}, 2025.

\bibitem{tum}
J.~Sturm, N.~Engelhard, F.~Endres, W.~Burgard, and D.~Cremers, ``A benchmark for the evaluation of rgb-d slam systems,'' in \emph{2012 IEEE/RSJ international conference on intelligent robots and systems}.\hskip 1em plus 0.5em minus 0.4em\relax IEEE, 2012, pp. 573--580.

\bibitem{bonn}
E.~Palazzolo, J.~Behley, P.~Lottes, P.~Giguere, and C.~Stachniss, ``Refusion: 3d reconstruction in dynamic environments for rgb-d cameras exploiting residuals,'' in \emph{2019 IEEE/RSJ International Conference on Intelligent Robots and Systems (IROS)}.\hskip 1em plus 0.5em minus 0.4em\relax IEEE, 2019, pp. 7855--7862.

\bibitem{davis}
F.~Perazzi, J.~Pont-Tuset, B.~McWilliams, L.~Van~Gool, M.~Gross, and A.~Sorkine-Hornung, ``A benchmark dataset and evaluation methodology for video object segmentation,'' in \emph{Proceedings of the IEEE conference on computer vision and pattern recognition}, 2016, pp. 724--732.

\end{thebibliography}
}

\end{document}